%% file: 0emnlp2021.tex
\pdfoutput=1

\documentclass[11pt]{article}

\usepackage[]{emnlp2021}

\usepackage{times}
\usepackage{latexsym}

\usepackage[T1]{fontenc}

\usepackage[utf8]{inputenc}

\usepackage{microtype}

\usepackage{graphicx}
\usepackage{natbib}
\usepackage{caption}

\usepackage{CJK}
\usepackage{tabularx}
\usepackage{bm}
\usepackage{amssymb}
\usepackage{booktabs}
\usepackage{url}
\usepackage{multicol}
\usepackage{color}
\usepackage{arydshln}
\usepackage{multirow}
\usepackage{amsmath} 
\usepackage{amsfonts}
\usepackage{subfigure}
\usepackage{verbatim}
\usepackage{bm}
\usepackage{tablefootnote}
\usepackage{natbib}

\usepackage{enumitem}
\setenumerate[1]{itemsep=0pt,partopsep=0pt,parsep=\parskip,topsep=0pt}
\setitemize[1]{itemsep=5pt,partopsep=5pt,parsep=\parskip,topsep=5pt}
\setdescription{itemsep=0pt,partopsep=0pt,parsep=\parskip,topsep=0pt}

\title{FCM: A Fine-grained Comparison Model for \\ Multi-turn Dialogue Reasoning}

\newcommand{\tabincell}[2]{\begin{tabular}{@{}#1@{}}#2\end{tabular}} 

\author{Xu Wang\textsuperscript{\rm 1}\footnotemark[1], Hainan Zhang\textsuperscript{\rm 2}\footnotemark[2], Shuai Zhao\textsuperscript{\rm 1}\footnotemark[2], \\ {\bf Yanyan Zou\textsuperscript{\rm 2}}, {\bf Hongshen Chen\textsuperscript{\rm 2}}, {\bf Zhuoye Ding\textsuperscript{\rm 2}}, {\bf Bo Cheng\textsuperscript{\rm 1}}, {\bf Yanyan Lan\textsuperscript{\rm 3}} \\
\textsuperscript{\rm 1}Beijing University of Posts and Telecommunications, Beijing, China\\ 
 \textsuperscript{\rm 2} Data Science Lab, JD.com, China \\  \textsuperscript{\rm 3}Institute for AI Industry Research, Tsinghua University, Beijing, China\\
 wxx@bupt.edu.cn, zhanghainan1990@163.com, \{zhaoshuaiby, chengbo\}@bupt.edu.cn, \\ \{zouyanyan6, dingzhuoye\}@jd.com, ac@chenhongshen.com, lanyanyan@tsinghua.edu.cn
}

\begin{document}
\maketitle
\renewcommand{\thefootnote}{\fnsymbol{footnote}}
\footnotetext[1]{Work done at Data Science Lab, JD.com.}
\footnotetext[2]{Corresponding Authors.}

\begin{abstract}
Despite the success of neural dialogue systems in achieving high performance on the leader-board, they cannot meet users' requirements in practice, due to their poor reasoning skills. The underlying reason is that most neural dialogue models only capture the syntactic and semantic information, but fail to model the logical consistency between the dialogue history and the generated response. 
Recently, a new multi-turn dialogue reasoning task has been proposed, to facilitate dialogue reasoning research. 
However, this task is challenging, because there are only slight differences between the illogical response and the dialogue history. 
How to effectively solve this challenge is still worth exploring.
This paper proposes a Fine-grained Comparison Model (FCM) to tackle this problem. Inspired by human's behavior in reading comprehension, a comparison mechanism is proposed to focus on the fine-grained differences in the representation of each response candidate. 
Specifically, each candidate representation is compared with the whole history to obtain a history consistency representation. 
Furthermore, the consistency signals between each candidate and the speaker's own history are considered to drive a model to prefer a candidate that is logically consistent with the speaker's history logic.
Finally, the above consistency representations are employed to output a ranking list of the candidate responses for multi-turn dialogue reasoning. Experimental results on two public dialogue datasets show that our method obtains higher ranking scores than the baseline models.

\end{abstract}

\input{1introduction}
\input{2related}

\input{3model}

\input{4experiments}

\input{5conclusion}

\bibliography{anthology}
\bibliographystyle{acl_natbib}

\end{document}

%% file: 1introduction.tex
\section{Introduction}
Nowadays, the neural dialogue system has achieved high performance and been widely studied in both industry \cite{8592630} and academia \cite{DBLP:journals/corr/abs-1906-00500}. 
However, the selected response often contradicts with the dialogue history, such as ``\textit{I am a teacher}'' as context but ``\textit{I work in the factory}'' in the response, which greatly affects the user experience. 
The underlying reason is that existing neural dialogue systems only model the syntactic and semantic relevance but fail to capture the logical consistency between the dialogue history and the generated response. 

\begin{table}[!t]
    \begin{centering}
    \scalebox{0.75}{
        \begin{tabular}{ll}
        \toprule
              speaker A: & Excuse me. How much is this suit?  \\
              \hdashline
              speaker B: & \tabincell{l}{It's \$750 today.} \\
              \hdashline
              speaker A: & \tabincell{l}{Wow, that is pretty expensive! }\\
              \hdashline
              speaker B: & \tabincell{l}{The material is \textit{\textbf{\textcolor{orange}{imported from Italy}}}. If you buy \\a suit with same material, it may be \$2000.} \\
              \hdashline
              speaker A: & \tabincell{l}{Uh-hah. But \textcolor{blue}{ I } \textit{\textbf{\textcolor{blue}{saw a suit just like this one}}},  \\and it was \$600. I still thought  it was expensive.}  \\
            \hline
             \multirow{6}*{\tabincell{l}{Candidates\\speaker B:}}& 1. \tabincell{l}{\textit{\textbf{\textcolor{blue}{ No suit has the style as it}}}. It's the style that \\makes it special.}\\
             &  2. \tabincell{l}{The material of this suit is \textit{\textbf{\textcolor{orange}{ imported from}} } \\ \textit{\textbf{\textcolor{orange}{France.}} }It makes the suit special.} \\ 
              & 3.  But \textit{\textbf{\textcolor{blue}{the color of our suit is very special.} }} \\
              & \textcolor{red}{ 4. \tabincell{l}{Although the suit you saw is same as it, the\\ material of our suit is \textit{\textbf{imported from Italy.}}}} \\
            \bottomrule
        \end{tabular}
        }
    \end{centering}
    \caption{\label{tb:example} An example of dialogue reasoning. The logical contradictions are labeled in the same color. The most proper and logically correct response is option 4, labeled in red.}
	\vspace{-5mm}
\end{table}

Recently, a new multi-turn dialogue reasoning task \cite{DBLP:conf/acl/CuiWLZZ20} has been proposed to facilitate conversation reasoning research. The goal of the dialogue reasoning task is to select the logical response from the extremely similar candidate responses. However, this task is challenging, because there are only slight differences between the illogical response and the dialogue history. For example in Table~\ref{tb:example}, option 1 and option 3 are in conflict with speaker A who has seen the same suit, and option 2 is in conflict with the imported country. Since the candidate options are only slightly different to the context, traditional dialogue models might tend to select semantically relevant yet illogical candidates, yielding incorrect responses.

As we all know, humans have the ability to make effective and efficient reasoning, because they usually focus on perceiving fine-grained details and compare the candidates at multiple-granularity levels accordingly. Taking Table~\ref{tb:example} as an example, by comparing option 2 and option 4, human can identify that the key difference is ``\textit{Italy}'' and ``\textit{France}'', which can be used as key information to distinguish 
such two options. Inspired by the human behaviors in reasoning, the fine-grained comparison between response and history should be introduced to improve the reasoning ability for the dialogue reasoning model.


Furthermore, the dialogue history from the same speaker is also critical for modeling the logical consistency~\cite{DBLP:journals/sigkdd/ChenLYT17}.  It is natural that a person holds his logical consistency when speaking. Thus the logical errors for the same speaker might hamper the user experience. 
 For example in Table~\ref{tb:example}, speaker B says the suit's material is imported from Italy in the dialogue history, but option 2 for him says the material is imported from France, which is a more obvious and serious logical error. Therefore, a reasoning model needs to consider the speaker's own historical consistency to distinguish logically incorrect candidate responses.

Inspired by the above analysis, we propose a Fine-grained Comparison Model (FCM) to improve the performance of multi-turn dialogue reasoning. To be specific, we firstly propose a comparison mechanism to compare every candidate response with all other ones. Secondly, we compare each candidate representation with the whole history and the speaker's own history to obtain the history and the speaker's consistency representations, respectively. Finally, we utilize the above consistency representations to output a ranking list of the candidate responses for multi-turn dialogue reasoning.

In our experiments, we utilize two public multi-turn dialogue datasets, named MuTual \cite{DBLP:conf/acl/CuiWLZZ20} and Ubuntu \cite{DBLP:conf/sigdial/LowePSP15}, to evaluate our proposed models. The results show that FCM has the ability to rank the candidate responses more accurately than the baseline models. We also conduct some case studies to demonstrate the superiority and soundness of FCM.

The main contributions of this paper include:
\begin{itemize}
\setlength{\itemsep}{0pt}
\setlength{\parsep}{0pt}
\setlength{\parskip}{0pt}
\item We introduce the response comparison mechanism to enable the dialogue model (e.g., BERT\cite{DBLP:conf/naacl/DevlinCLT19}) to have fine-grained detail perception ability, which tackles the difficulty of subtle differences between candidates and dialogue history in dialogue reasoning.
\item We model the speaker's own logical consistency to further enhance the reasoning ability for dialogue reasoning task.
\item We experiment on two public multi-turn dialogue datasets to demonstrate the effectiveness of our proposed model FCM.
\end{itemize}

%% file: 2related.tex
\section{Related Work}
Recently, multi-turn dialogue has gained more attention in both industry \cite{DBLP:conf/acl/WuLZZW20,DBLP:conf/naacl/ZhanZCDBL21} and academia \cite{DBLP:conf/acl/ChoM20},  compared with single-turn dialogue \cite{DBLP:conf/coling/MouSYL0J16,DBLP:conf/ijcai/ZhangLGXC18,DBLP:conf/emnlp/LiMSJRJ17}. \citet{SERBAN:HRED1} proposes a hierarchical recurrent encoder-decoder (HRED) model which uses the hierarchical encoder-decoder framework to model the relevance of the context and response.
\citet{DBLP:conf/acl/WuWXZL17} uses HRED to model relationships among utterances to enhance the performance of the retrieval-based chatbot.
\citet{DBLP:conf/www/ChenRTZY18} adds the hierarchical structure and the variable memory network into a
neural encoder-decoder network, which can capture both the high-level abstract variations and long-term memories during dialogue tracking. \citet{DBLP:conf/coling/ZhangCWZLZL18} adopts dynamic and static attention to weigh the importance of each utterance in a conversation and then obtain the contextual representation.
\citet{DBLP:conf/acl/ZhangLPGC19} utilizes hierarchical self-attention mechanism to solve the position bias problem of dialogue models.

Although these models have achieved good performance on datasets like DailyDialog \cite{DBLP:conf/ijcnlp/LiSSLCN17} and ECD \cite{DBLP:conf/coling/ZhangLZZL18}, there is still a giant gap between high performance on the leader-board and poor practical user experience \cite{DBLP:conf/acl/CuiWLZZ20}. These models frequently generate responses that are logically incorrect.
One possible reason is that the previous models \cite{DBLP:conf/aaai/YoungCCZBH18,DBLP:journals/corr/abs-1906-04341} only solve the cases through linguistic information matching yet lack of logical reasoning. 

Obviously, a dialogue reasoning dataset that can help the model to detect the illogical response is extremely necessary. Recently, an open domain multi-turn dialogue reasoning dataset (MuTual) \cite{DBLP:conf/acl/CuiWLZZ20} is proposed to facilitate the reasoning capabilities of conversation models.
In particular, given a context, it prepares four candidate responses and all of them are relevant to the context, but singly one has the correct logic. It requires fine-grained reasoning ability between context and response to make the correct choice.
Current multi-turn dialogue models \cite{DBLP:conf/naacl/DevlinCLT19,DBLP:conf/coling/ZhangLZZL18,DBLP:journals/corr/abs-1907-11692,DBLP:conf/acl/WuWXZL17}, which perform well on existing benchmarks, have declined on this dataset, which proves that the reasoning ability of these models is still insufficient. Therefore, how to enhance the logical reasoning ability of dialogue models is worth discussing.

Logic consistency of history is essential in human-like behavior. For the multi-choice reasoning task \cite{DBLP:conf/aaai/ZhuWQL18}, there are only slight differences between the illogical response and the dialogue history, so a fine-grained comparison mechanism is required to infer the logic between each candidate response and the dialogue history. Previous models \cite{DBLP:conf/aaai/YoungCCZBH18,DBLP:conf/naacl/DevlinCLT19} ignore the modeling of fine-grained comparison, which leads to the loss of reasoning ability of their models. In this paper, we overcome this challenge and propose a fine-grained comparison mechanism to perform history consistency and speaker consistency respectively.

\begin{figure*}[htbp!]
	\centering  
	\small
	\includegraphics[width=0.96\linewidth]{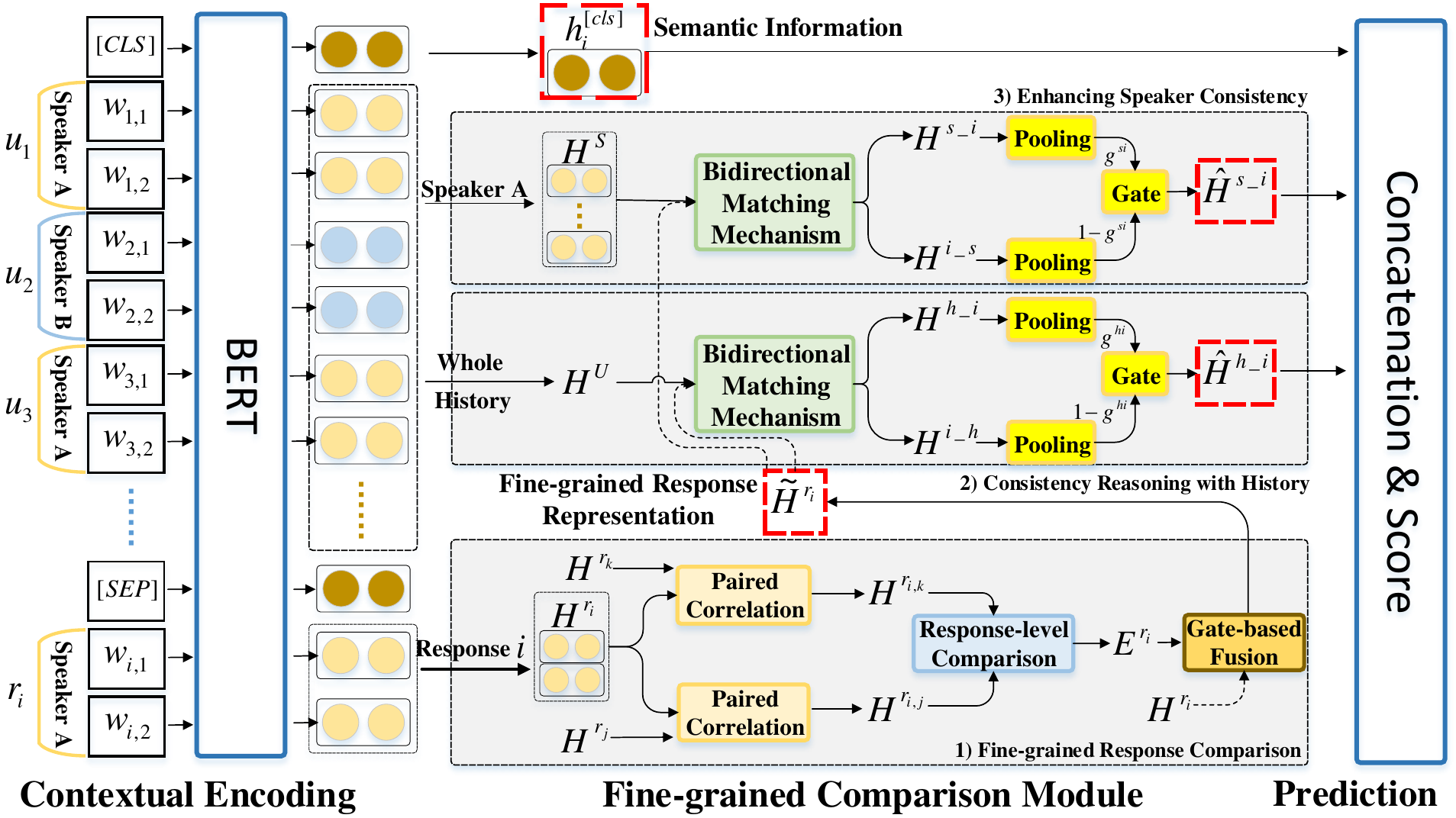}  
	\caption{The proposed FCM network for multi-turn dialogue reasoning task. It contains three components: 1) Contextual Encoding, 2) Fine-grained Comparison Module, and 3) Response Prediction.}
	\label{fig:architecture}   
		\vspace{-3mm}
\end{figure*}

%% file: 3model.tex
\section{Model}
In this section, we first describe the task definition and then introduce our FCM model in detail, with the architecture shown in Figure~\ref{fig:architecture}. Our FCM model consists of three components, i.e., Contextual Encoding, Fine-grained Comparison Module, and Response Prediction.
Firstly, we utilize the pre-trained language model BERT to encode each token of context and response into a fixed-length vector, which carries contextual information.
Secondly, we utilize our comparison mechanism to obtain the fine-grained response representation and then compare such obtained representation with both the whole history and the speaker's own history.
Finally, we fuse consistency representation and semantic information, and then use a linear layer to obtain the candidate response score for the multi-choice prediction process.

\subsection{Task Definition}
Given a dialogue context $U = \{u_1,u_2,...,u_N\}$ and a candidate response set $R = \{r_1,r_2,...,r_M\}$, where $u_i = \{w^{u}_{i,1},w^{u}_{i,2}, ...,w^{u}_{i,l_i}\}$ is an utterance with $l_i$ tokens, and $r_i = \{w^{r}_{i,1},w^{r}_{i,2}, \noindent ...,w^{r}_{i,j_i}\}$ is a candidate response with $j_i$ tokens, the goal of this task is to select the proper and logical response based on the conditional probability distribution, i.e., $P(r_i|U,R)$, where $r_i \in R$.

\subsection{Contextual Encoding} \label{Contextual Encoding}
Given each input pair $(U,r_i)$, 
we concatenate the context and each candidate and then feed them into the pre-trained BERT to obtain the fixed-length vector of each token in the context and response, which is denoted as:
\begin{equation}
    [H^{U}; H^{r_i}] = BERT( <U; r_i> ),    \label{formual1}
\end{equation}
where $BERT(\cdot)$ returns the last layer output of the encoder.
$<;>$ means concatenation of two sequences. 
$H^{U} \in \mathbb{R}^{|U| \times d}$ and $H^{r_i} \in \mathbb{R}^{|r_i| \times d}$ are the token-level vectors of context $U$ and  candidate $r_i$, respectively. $d$ is the dimension of the hidden state. 
Besides, we obtain the summary vector $h^{[cls]}_{i} \in \mathbb{R}^d$ for input pair $(U,r_i)$, which carries the semantic information of the whole input \cite{DBLP:conf/naacl/DevlinCLT19}.

\subsection{Fine-grained Comparison Module}
In this section, we introduce our fine-grained comparison module, including three steps: fine-grained response comparison, consistency reasoning with history, and enhancing speaker consistency. Specifically, a fine-grained response comparison mechanism is firstly utilized to imitate human behaviors, which aims to compare the correlation and difference between candidate responses at multi-granularity levels. Then, the history-aware bidirectional matching method is utilized to infer the fine-grained logical consistency between the candidate $r_i$ and the whole history $U$. Finally, another bidirectional matching model is used to infer the speaker consistency between the response $r_i$ and the speaker's own history.

\subsubsection{Fine-grained Response Comparison}\label{sec:compare}
For each response $r_i$, a fine-grained attention mechanism is used to compare it with all other responses to get the fine-grained comparison information.

\paragraph{Paired Correlation} 
Given the hidden vectors $H^{r_i}$ and $H^{r_j}$, we calculate the word-level attention between them to obtain the similarity matrix $A^{r_{i,j}}$, which is defined as:
\begin{equation}
\label{formual2}
    \begin{split}
        & A^{r_{i,j}} = \bigg [\frac{\exp(a^{r_{i,j}}_{mn})}{\sum_{n}\exp(a^{r_{i,j}}_{mn})}\bigg ]_{m,n},\\
        & a^{r_{i,j}}_{mn} = W_{1}^{T} [H^{r_i}_m;H^{r_j}_n;H^{r_i}_m \odot H^{r_j}_n],  \\
    \end{split}
\end{equation}
where $\odot$ is the element multiplication between two matrices. $H^{r_i}_m$ is the hidden representation of $m$th token in $r_i$, and $W_{1}^{T} \in \mathbb{R}^{1 \times 3d}$ is a learned parameter. $a^{r_{i,j}}_{mn}$ means the similarity between the $m$th word of $r_i$ and the $n$th word of $r_j$. 

Given the similarity matrix $A^{r_{i,j}}$, the paired correlation information $H^{r_{i,j}}$ is defined as:
\begin{equation}
    \label{formual3}
    \begin{split}
    & H^{r_{i,j}} = [H^{r_i} - \overline{H}^{r_{i,j}}; H^{r_i} \odot \overline{H}^{r_{i,j}}],\\
    &\overline{H}^{r_{i,j}} = A^{r_{i,j}} H^{r_j},   \\
    \end{split}
\end{equation}
where $\overline{H}^{r_{i,j}}$ highlights the similar part of $r_j$ with $r_i$, and $H^{r_{i,j}}$ represents the different part between the candidate $r_i$ and $r_j$.

\paragraph{Response-level Comparison} Given the correlation information $H^{r_{i,j}}$ for $r_i$, the response-level compared information $E^{r_i}$ is defined as:
\begin{equation}
    E^{r_i} = tanh \bigg (  \bigg [  \{H^{r_{i,j}} \}_{j  \neq i}   \bigg ]  W_{2} + b_{2} \bigg ),   \label{formual4}
\end{equation}
where $W_{2} \in \mathbb{R}^{ 2d(M - 1) \times d }$ and $b_{2} \in \mathbb{R}^{d}$ are learned parameters. $M$ refers to the total number of candidate responses.

\paragraph{Gate-based Fusion} Given the contextual encoding features $H^{r_i}$ and the response-level compared information $E^{r_i}$ of response $r_i$, an element-wise gating mechanism is utilized to obtain the fine-grained response representation $\widetilde{H}^{r_i}$, which is defined as:
\begin{equation}
\label{formual5}
\begin{split}
    &g^{r_i} = \sigma( [E^{r_i}; H^{r_i}] W_{3} + b_{3} ),\\
    &\widetilde{H}^{r_i} = g^{r_i} \odot E^{r_i} + (1-g^{r_i}) \odot H^{r_i},
\end{split}
\end{equation}
where $W_{3} \in \mathbb{R}^{2d \times d}$ and $b_{3} \in \mathbb{R}^{d}$ are learned parameters. $g^{r_i} \in \mathbb{R}^{|r_i| \times d}$ stands for the element-wise gate value. 

\subsubsection{Consistency Reasoning with History}\label{sec:matching}
Given the context representation $H^{U} \in \mathbb{R}^{|U| \times d}$ and the fine-grained response representation $\widetilde{H}^{r_i} \in \mathbb{R}^{|r_i| \times d}$, we design a bidirectional matching mechanism to obtain the response-aware history representation $H^{h\_i}$ and history-aware response representation $H^{i\_h}$ respectively, which is defined as:
\begin{equation}
    \begin{split}
         & A^{h\_i} = SoftMax(H^{U}W_{4}\widetilde{H}^{{r_i}^{T}}), \\
         & A^{i\_h} = SoftMax(\widetilde{H}^{r_i} W_{5} {H^{U}}^{T}), \\
         & H^{h\_i} = Relu(A^{h\_i} \widetilde{H}^{r_i} W_{6}), \\
         & H^{i\_h} = Relu(A^{i\_h} H^{U} W_{7}), \\
    \end{split}
    \label{formual6}
\end{equation}
where $W_{4}$, $W_{5}$, $W_{6}$, and $W_{7} \in \mathbb{R}^{d \times d}$ are learned parameters.
$A^{h\_i}$ and $A^{i\_h}$ are the word-level attention matrices between the whole history $U$ and response $r_i$, which focus on different perspectives.

Given the two representations $H^{i\_h}$ and $H^{h\_i}$ as input, we utilize a gate mechanism to fuse them to get the history-aware consistency information $\hat{H}^{h\_i}$, which is defined as:
\begin{equation}
    \begin{split}
        & E^{h\_i} = MaxPooling(H^{h\_i}), \\
        & E^{i\_h} = MaxPooling(H^{i\_h}), \\
        & g^{hi} = \sigma (E^{h\_i}W_{8} + E^{i\_h}W_{9} + b_{4}), \\
        & \hat{H}^{h\_i} =g^{hi} \odot E^{h\_i} + (1-g^{hi}) \odot E^{i\_h},
    \label{formual7}
    \end{split}
\end{equation}
where $W_{8}$, $W_{9} \in \mathbb{R}^{d \times d}$  and $b_{4} \in \mathbb{R}^{d}$ are learned parameters. $MaxPooling$ means row-wise max pooling operation.
$g^{hi}$ means the element-wise gate value.

\subsubsection{Enhancing Speaker Consistency}
Given the speaker's own context $H^{S}$ and the compared response representation $\widetilde{H}^{r_i}$ of $r_i$ from section~\ref{sec:compare}, we utilize another bidirectional matching module to obtain the response-aware speaker representation $H^{s\_i}$ and speaker-aware response representation $H^{i\_s}$, respectively, which is with the similar definition in section~\ref{sec:matching}:
\begin{equation}
    \begin{split}
         & A^{s\_i} = SoftMax(H^{S}W_{10}\widetilde{H}^{{r_i}^{T}}), \\
         & A^{i\_s} = SoftMax(\widetilde{H}^{r_i} W_{11} {H^{S}}^{T}), \\
         & H^{s\_i} = Relu(A^{s\_i} \widetilde{H}^{r_i} W_{12}), \\
         & H^{i\_s} = Relu(A^{i\_s} H^{S} W_{13}), \\
    \end{split}
    \label{formual8}
\end{equation}
where $A^{s\_i}$ and $A^{i\_s}$ are the word-level attention matrices between the speaker's own context $S$ and response $r_i$.

Given the two representations $H^{i\_s}$ and $H^{s\_i}$ as input, the speaker-aware consistency information $\hat{H}^{s\_i}$ is defined as:
\begin{equation}
    \begin{split}
        & E^{s\_i} = MaxPooling(H^{s\_i}), \\
        & E^{i\_s} = MaxPooling(H^{i\_s}), \\
        & g^{si} = \sigma (E^{s\_i}W_{14} + E^{i\_s}W_{15} + b_{5}), \\
        & \hat{H}^{s\_i} =g^{si} \odot E^{s\_i} + (1-g^{si}) \odot E^{i\_s}.
    \label{formual9}
    \end{split}
\end{equation}

\subsection{Response Prediction}

Given the semantic information $h^{[cls]}_{i}$, history-aware consistency $\hat{H}^{h\_i}$ and speaker-aware consistency $\hat{H}^{s\_i}$, we concatenate them to get the reasoning information $H^{i}$ for response $r_i$, which is defined as:
\begin{equation}
    H^{i} = [h^{[cls]}_{i} ;\hat{H}^{h\_i}; \hat{H}^{s\_i} ].  \label{prediction}
\end{equation}

The score $P(r_i|U, R)$ of the $i$th candidate response is computed as follows:
\begin{equation}
    P(r_i|U, R ) = \frac{exp(W_{16} H^{i} + b_{6})}{\sum^{M}_{i=0} exp(W_{16} H^{i} + b_{6}) },
\end{equation}
where $W_{16} \in \mathbb{R}^{1 \times 3d}$ and $b_{6} \in \mathbb{R}^{1}$ are learned parameters.

The loss function is defined as:
\begin{equation}
    J(\theta) = - \frac{1}{N}\sum  {\rm log}P(\hat{r_i}|U, R ) + \lambda||\theta||^{2}_2,
\end{equation}
where $\lambda$ is a hyperparameter, $\theta$ are all trainable parameters, $N$ is the size of training examples in the dataset, and $\hat{r_i}$ is the ground-truth response.

%% file: 4experiments.tex
\section{Experiments}
In this section, we conduct experiments on MuTual reasoning and Ubuntu dialogue datasets to evaluate our proposed method.
\subsection{Experimental Settings}
We first introduce some empirical settings, including datasets, baseline methods, parameter settings, and evaluation measures.
\subsubsection{Datasets}
We test our model on two public multi-turn dialogue datasets: MuTual and Ubuntu.

MuTual \cite{DBLP:conf/acl/CuiWLZZ20} consists of 8860 manually annotated dialogues, which is based on Chinese student English listening comprehension exams. Each dialogue has two speakers speaking in turn and contains four candidate responses. The goal of this task is to select the correct and logical response according to the historical contexts. 
The training, validation, and testing sets contain 7088, 886, and 886 pairs, respectively.

Ubuntu \cite{DBLP:conf/sigdial/LowePSP15} consists of English multi-turn dialogues about technical support collected from chat logs on Ubuntu forum. 
The dataset contains 1 million context-response training pairs, 0.5 million validation pairs, and 0.5 million testing pairs. Each pair has one positive response and nine negative responses. Because there are some sessions in Ubuntu dataset that require reasoning, we utilize Ubuntu dataset to verify the reasoning ability of FCM.

\begin{table*}
    \begin{center}
        \scalebox{0.9}{
        \begin{tabular}{lcccccc}
            \toprule
             \multirow{2}*{Model}& \multicolumn{3}{c}{Dev} & \multicolumn{3}{c}{Test} \\
             & $\mathbf{R_{4}@1}$ & $\mathbf{R_{4}@2}$ & $\mathbf{MRR}$ & $\mathbf{R_{4}@1}$ & $\mathbf{R_{4}@2}$ & $\mathbf{MRR}$ \\
            \hline
            TF-IDF \cite{DBLP:conf/sigir/Paik13} & 0.276 & 0.541 & 0.541 & 0.279 & 0.536 & 0.542 \\
            
            Dual LSTM \cite{DBLP:conf/sigdial/LowePSP15} & 0.266 & 0.528 & 0.538 & 0.260 & 0.491 & 0.743 \\
            
            SMN \cite{DBLP:conf/acl/WuWXZL17} & 0.274 & 0.524 & 0.575 & 0.299 & 0.585 & 0.595 \\

            DAM \cite{DBLP:conf/acl/WuLCZDYZL18} & 0.239 & 0.463&0.575 & 0.241 & 0.465 & 0.518 \\
            \hdashline
            GPT-2\cite{radford2019language} &0.335&0.595&0.586&0.332&0.602&0.584 \\
            BERT \cite{DBLP:conf/naacl/DevlinCLT19} & 0.657 & 0.867 & 0.803 & 0.648 & 0.847 & 0.795 \\
            BERT-MC \cite{DBLP:conf/naacl/DevlinCLT19} & 0.661 &0.871 &0.806 &0.667 & 0.878 &0.810 \\
            \hdashline
            \textbf{FCM} & \textbf{0.696} & \textbf{0.884} & \textbf{0.824} & \textbf{0.692} & \textbf{0.884} & \textbf{0.823} \\
            \bottomrule
            
        \end{tabular}
        }
    \end{center}
    \caption{\label{experiments result} The metric-based evaluation on MuTual dataset.}
\end{table*}

\subsubsection{Baseline Methods}
We use 11 baselines for comparison, including the traditional TF-IDF \cite{DBLP:conf/sigir/Paik13}, Dual LSTM \cite{DBLP:conf/sigdial/LowePSP15}, SMN \cite{DBLP:conf/acl/WuWXZL17}, DAM \cite{DBLP:conf/acl/WuLCZDYZL18}, BERT \cite{DBLP:conf/naacl/DevlinCLT19}, BERT-MC~\cite{DBLP:conf/acl/CuiWLZZ20}, GPT-2 \cite{radford2019language}, Deep Utterance Aggregation (DUA) \cite{DBLP:conf/coling/ZhangLZZL18}, Interaction-over-Interaction (IoI) \cite{DBLP:conf/acl/TaoWXHZY19}, Multi-hop Selector (MSN) \cite{DBLP:conf/emnlp/YuanZLLZHH19} and Multi-Representation Fusion (MRFN) \cite{DBLP:conf/wsdm/TaoWXHZY19}.

\subsubsection{Parameter Settings}
We utilize the open-source pre-trained model BERT$_\text{base}$\footnote{\href{https://github.com/huggingface/transformers}{https://github.com/huggingface/transformers}} for the dialogue reasoning task. 
BERT$_\text{base}$ has 12-layer transformer blocks, 768 hidden-size, and 12 self-attention heads. It totally contains 110M parameters. 
In order to make a fair comparison between our model and baselines, 1) for MuTual dataset, we refer to \cite{DBLP:conf/acl/CuiWLZZ20} and set the max input sequence length to 350. We set the dropout rate to 0.2. The $L2$ weight decays $\lambda$ is set to 0.01.
We employ Adam \cite{DBLP:journals/corr/KingmaB14} to optimize the model with a learning rate 1e-6.
We run the experiments on two TITAN XP GPUs with 12G memory and train for 10 epochs with batch size of 4; 2) for Ubuntu dataset, we use the same evaluation metrics which are used in previous works \cite{DBLP:conf/cikm/GuLLLSWZ20,DBLP:conf/coling/ZhangLZZL18}.

\subsubsection{Evaluation Measures}

We consider the dialogue reasoning task as a retrieval-based response selection task and apply traditional information retrieval metrics. On MuTual, we display the recall and Mean Reciprocal Rank measures\cite{DBLP:conf/lrec/VoorheesT00}, i.e., $\mathbf{R_{4}@1}$, $\mathbf{R_{4}@2}$, and $\mathbf{MRR}$. On Ubuntu, we use $\mathbf{R_{10}@1}$, $\mathbf{R_{10}@2}$, and  $\mathbf{R_{10}@5}$ for evaluation.

\begin{table}[!t]
    \begin{center}
    \scalebox{0.78}{
        \begin{tabular}{lccc}
        
            \toprule
            Model & $\mathbf{R_{10}@1}$ & $\mathbf{R_{10}@2}$ & $\mathbf{R_{10}@5}$  \\
            \hline
            SMN \cite{DBLP:conf/acl/WuWXZL17} & 0.726 & 0.847 & 0.961  \\
            DUA \cite{DBLP:conf/coling/ZhangLZZL18} & 0.752 & 0.868 & 0.962  \\
            DAM \cite{DBLP:conf/acl/WuLCZDYZL18}& 0.767 & 0.874 & 0.969  \\
            IoI \cite{DBLP:conf/acl/TaoWXHZY19}& 0.796 & 0.894 & 0.974  \\
            MSN \cite{DBLP:conf/emnlp/YuanZLLZHH19}& 0.800 & 0.899 & 0.978  \\
            MRFN \cite{DBLP:conf/wsdm/TaoWXHZY19}&0.786 & 0.886 & 0.976  \\
            BERT \cite{DBLP:conf/naacl/DevlinCLT19}& 0.808 & 0.897 & 0.975 \\
            \hdashline
            \textbf{FCM} & \textbf{0.816} & \textbf{0.908} & \textbf{0.983}\\
            \bottomrule
        \end{tabular}
        }
    \end{center}
    \caption{\label{Ubuntu}The metric-based evaluation on Ubuntu.}
    \vspace{-0mm}
\end{table}

\subsection{Experimental Results}
Now we demonstrate our experimental results on
the two public datasets.
\subsubsection{Metric-based Evaluation}
The metric-based evaluation results on MuTual and Ubuntu are shown in Table~\ref{experiments result} and Table~\ref{Ubuntu}.
From the results, we can see that the performance of well-designed RNN-based networks, such as Dual LSTM and SMN, is relatively poor, which demonstrates that such traditional models cannot deal with the dialogue reasoning task. Although the performance of BERT is better than other baseline models, merely using self-attention between context and candidate responses still misses fine-grained consistency information.
With the introduction of fine-grained comparison information, our FCM model outperforms all baseline models. Take the $\mathbf{R_{4}@1}$ and $\mathbf{R_{4}@2}$ on the MuTual dev set as an example, the $\mathbf{R_{4}@1}$ and $\mathbf{MRR}$ of our FCM model are 69.6\% and 82.4\%, respectively, which is significantly better than that of BERT-MC, i.e., 3.5\% and 1.8\%. 
On Ubuntu, we find that FCM also outperforms the BERT model on three metrics.
In summary, our FCM model has the ability to select a more logically consistent response than baselines.

\subsubsection{Case Study}
To facilitate a better understanding of our model, we present the examples on MuTual in Table~\ref{Case Study}. From the results, we can see that our model is more accurate in multi-choice prediction than the traditional baselines and transformer models. In this example, both response 1 and response 4 state that ``\textit{Sichuan food}'' is speaker B's favorite food. BERT and BERT-MC agree with this statement. However, from speaker A's historical utterance  ``\textit{I know Guangdong food is your favorite kind of Chinese food.}'', we can guess that speaker B's favorite is ``\textit{Guangdong food}''.
According to the above analysis, we argue that the two baselines lack the modeling of fine-grained consistency reasoning, especially for the speaker's own consistency,  so they predict the wrong response, which proves the effectiveness of logical consistency.

Given the wrong response 2 and the correct response 3, the biggest difference between them is the utterance ``\textit{I know where is it.}''. 
Using the consistency reasoning mechanism, we can infer from the speaker A's historical utterance ``\textit{I do not know where it is.}'' that this statement is wrong. However, baselines do not use the fine-grained response comparison mechanism and can not focus on the fine-grained differences between response 2 and response 3, and then it predicts the wrong response 2, which illustrates the importance of fine-grained response comparison in logic consistency. In summary, compared with baseline models, our proposed model FCM, which carries the fine-grained comparison ability, is capable of inferring logic consistency more accurately for the multi-turn dialogue reasoning task.

\begin{table}[!t]
    \begin{center}
        \scalebox{0.74}{
        \begin{tabular}{lccc}
         \multicolumn{4}{c}{MuTual Dev}\\
            \toprule
             Model& $\mathbf{R_{4}@1}$ & $\mathbf{R_{4}@2}$ & $\mathbf{MRR}$  \\
            \hline
            \textbf{FCM} & \textbf{0.696} & \textbf{0.884} & \textbf{0.824} \\
            \quad w/o Response Comparison & 0.680 & 0.875 & 0.815 \\
            \quad w/o Consistency with History & 0.690 & 0.878 & 0.820  \\
            \quad w/o Speaker Consistency & 0.685 & 0.880 & 0.816  \\
            \bottomrule
            \multicolumn{4}{c}{ Ubuntu} \\
            \toprule
             Model & $\mathbf{R_{10}@1}$ & $\mathbf{R_{10}@2}$ & $\mathbf{R_{10}@5}$ \\
            \hline
            \textbf{FCM} &  \textbf{0.816} & \textbf{0.908} & \textbf{0.983} \\
            \quad w/o Response Comparison  &  0.814 & 0.904 & 0.980  \\
            \quad w/o Consistency with History  &  0.812 & 0.903 & 0.978  \\
            \quad w/o Speaker Consistency &  0.812 & 0.902 & 0.979  \\
            \bottomrule
        \end{tabular}
        }
    \end{center}
    \caption{\label{Ablation Study}Ablation experimental results of our FCM model on MuTual Dev and Ubuntu datasets.}
    \vspace{0pt}
\end{table}

\subsection{Ablation Study} \label{Ablation}
To study the contributions of the main components in FCM, we conduct ablation experiments, mainly including removing the \textbf{Fine-grained Response Comparison} module, \textbf{Consistency Reasoning with History} module, and \textbf{Enhancing Speaker Consistency} module from our proposed model, respectively.

The results on MuTual and Ubuntu are shown in Table~\ref{Ablation Study}. We can see that without response comparison, the performance of the model drops in all three metrics. Taking the Mutual dev set as an example, the model decreased by 1.6\%, 0.9\%, and 0.9\% on $\mathbf{R_{4}@1}$, $\mathbf{R_{4}@2}$, and $\mathbf{MRR}$, respectively, which proves the importance of fine-grained response comparison in the reasoning process. 
When without whole history consistency reasoning, we can see that the performance is reduced, which demonstrates the necessity of modeling consistency between each candidate and the history. 
When without speaker consistency reasoning, we can see that the performance is also reduced. Taking the MuTual dev set as an example, the measures decreased by 1.1\% and  0.8\% on $\mathbf{R_{4}@1}$ and $\mathbf{MRR}$, respectively.
This reduction proves that enhanced speaker consistency is helpful for the model to calculate the score of each candidate response. 
\begin{table}[!t]
    \begin{center}
        \scalebox{1}{
        \begin{tabular}{lccc}
            \toprule
            Model & $\mathbf{R_{4}@1}$ & $\mathbf{R_{4}@2}$ & $\mathbf{MRR}$ \\
            \hline
            \textbf{FCM} & \textbf{0.6964} & \textbf{0.8841} & \textbf{0.8236}\\
            \quad coarse-grained & 0.6817 & 0.8772 & 0.8161\\
            \hdashline
            \quad simple-add & 0.6839 & 0.8837 & 0.8192 \\
            \quad  no-source & 0.6884 & 0.8818 & 0.8202 \\
            \quad  no-gate & 0.6871 & 0.8795 & 0.8189 \\
            \bottomrule
        \end{tabular}
        }
    \end{center}
    \caption{\label{Response Comparison}Analysis of fine-grained response comparison.}
	\vspace{-0mm}
\end{table}

\begin{table*}[!t]
    \begin{center}
    \scalebox{0.8}{
        \begin{tabular}{lll}
            \multicolumn{3}{c}{The Example of MuTual Dataset}\\
            \toprule
            \multirow{14}*{Context} & speaker A & It's already 30. How about preparing supper now? \\
            ~ & speaker B & But I don't want to cook today. I'm tired of cooking every day.  \\
            \cline{2-3}
            ~ & speaker A &  \tabincell{l}{How about having supper out tonight?  There is a new Chinese restaurant on  the third \\ street. Tom went there yesterday, and he said it was great.} \\
            ~ & speaker B & Really? What kinds of food does it have? You know, I don't like food that's too spicy. \\
            \cline{2-3}
            ~ & speaker A & \tabincell{l}{Don't worry. One of the chefs is from Guangdong. \\ \textit{\textbf{\textcolor{blue}{I know Guangdong food is your favorite kind of Chinese food}}}.} \\
            ~ & speaker B &\tabincell{l}{That's great.  Do you know how to get there?}  \\
            \cline{2-3}
            ~ & speaker A & \tabincell{l}{\textit{\textbf{\textcolor{orange}{I do not know where it is. I just know it's on the third street}}}. \\ Don't worry. I'm sure we will find it.}\\
            ~ & speaker B & But I don't feel like walking now. It's still so hot outside.\\
            \cline{2-3}
            ~ & speaker A & Then how about asking Tom to pick us up? We can treat him to supper.\\
            ~ & speaker B & That's a good idea.\\
            \hline
            \multirow{4}*{Candidates} &1) speaker A: & \tabincell{l}{So since \textit{\textbf{\textcolor{blue}{Sichuan food is your favorite kind}}} of Chinese food, \\ why don't we go there after work?} \\
            ~ &2) speaker A: & Ok, dear, let's go. \textit{\textbf{\textcolor{orange}{ I know where is it}}}. \\
            ~ & \textcolor{red}{\textbf{3) speaker A:}} & \textcolor{red}{\textbf{Ok, dear, let's go!} }\\
            ~ &4) speaker A: & Great. After school, we can go there to eat \textit{\textbf{\textcolor{blue}{your favorite Sichuan food}}}. \\
            \hline
            
            Model & BERT & BERT-MC \quad \quad \quad  GPT-2 \quad \quad \quad \textbf{FCM}\\
            
            \hline
            Predictions & \quad 1) & \quad \quad 4) \qquad \qquad \qquad 2) \quad  \quad \qquad \textbf{3)} \\
            \hline
        \end{tabular}
    }
    \end{center}
    \vspace{-0mm}
    \caption{\label{Case Study}The selected candidate response from our FCM model and baselines on MuTual.}
\end{table*}
\subsection{Analysis of Fine-grained Response Comparison}
To prove the effectiveness of our fine-grained response comparison module, we have conducted experiments on each operation of this module, mainly focusing on the operations of paired correlation calculation and gate-based fusion.

\paragraph{Analysis of paired correlation calculation}
To check the validity of these operations, we select a simple method to get the coarse-grained correlation, instead of Equation~\ref{formual2} and Equation~\ref{formual3}, which is defined as follows:
\begin{equation}
    \begin{split}
         & A^{r_{i,j}} = SoftMax(H^{r_i}(H^{r_j})^{T}), \\
         & H^{r_{i,j}} = A^{r_{i,j}}H^{r_j}.
    \end{split}
\end{equation}
From the results in Table~\ref{Response Comparison}, with (coarse-grained), we observe that the performance of the FCM decreases in $\mathbf{R_{4}@1}$, $\mathbf{R_{4}@2}$, and $\mathbf{MRR}$, which proves the effectiveness of our designed fine-grained method for paired correlation. 

\paragraph{Analysis of gate-based fusion}
In order to prove the effectiveness of the gate-based fusion, we design the following experiments: 1) (no-source): At the calculation of gate value $g^{r_i}$ in Equation~\ref{formual5}, we remove the original response representation $H^{r_i}$ and only utilize the updated response representation $E^{r_i}$; 2) (simple-add): In Equation~\ref{formual5}, we remove the operation of the weight-based summarization; 3) (no-gate): We directly remove the gating mechanism and use the output of Equation~\ref{formual4} to represent the response-level compared information.  

The results are shown in Table~\ref{Response Comparison}, and we obtain the following conclusions: 1) With (no-source), the performance of our model decreases, which shows that the retention of the original information is necessary to calculate gate value $g^{r_i}$; 2) With (simple-add), the updated response information and the original response information cannot be treated equally; 3) With (no-gate), we still get a lower performance, which means that after obtaining new knowledge, the response-level compared information may lose the original information.

\subsection{Generality of FCM}
\begin{figure}[t]
	\centering  
	\small
	\includegraphics[width=1\linewidth]{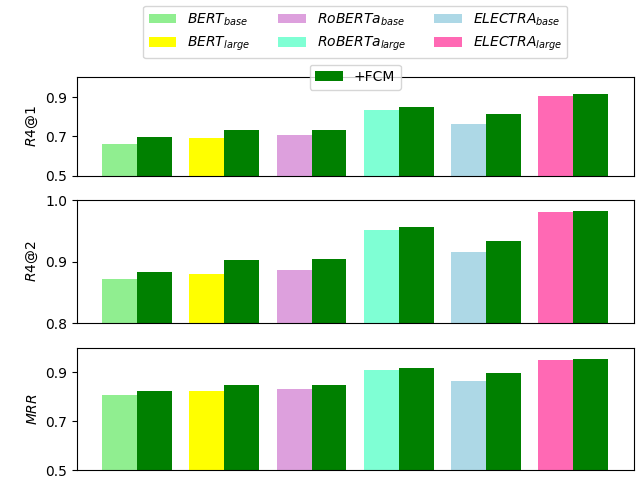}  
	\caption{The $\mathbf{R_{4}@1}$, $\mathbf{R_{4}@2}$ and $\mathbf{MRR}$ performance of different pre-trained language models with FCM on MuTual.
	}
	\label{Generality}   
	\vspace{-5mm}
\end{figure}

We test the generality of FCM in the pre-trained language models. Specifically, we apply FCM to the widely used models: $\text{BERT}_\text{base}$, $\text{BERT}_\text{large}$, $\text{RoBERTa}_\text{base}$, $\text{RoBERTa}_\text{large}$, $\text{ELECTRA}_\text{base}$ and $\text{ELECTRA}_\text{large}$, respectively.  The experiment results are shown in Figure~\ref{Generality}. From the results, we can discover that after applying our model to different pre-trained language models,  their performance can all be improved, which proves that FCM is generally effective to the widely used pre-trained language models.

%% file: 5conclusion.tex
\section{Conclusion}
In this paper, we focus on multi-turn dialogue reasoning tasks and propose the FCM model. The motivation comes from the fact that the widely used dialogue models only focus on the syntactic and semantic relevance but fail to model the logical consistency between the dialogue history and the generated response.
This task is challenging because there are only slight differences between the illogical response and the dialogue history. Our core idea is to propose a fine-grained comparison mechanism to focus on the fine-grained differences in the representation of each response candidate, and then each candidate representation is compared with the history to obtain a consistency score.
In the future, we plan to further investigate the fine-grained correlation between different speakers, and utilize this information to help improve our model.


\section*{Acknowledgements}
This work is supported by Beijing Nova Program of Science and Technology (Grant No.Z191100001119031),  National Key Research and Development Program of China (No. 2018YFB1003804), Guangxi Key Laboratory of Cryptography and Information Security (No.GCIS202111), the Open Program of Zhejiang Lab (Grant No.2019KE0AB03), the Beijing Academy of Artificial Intelligence (BAAI), and the National Natural Science Foundation of China (NSFC) (No.61773362). Xu Wang is supported by BUPT Excellent Ph.D. Students Foundation under grant CX2019137. 
We thank the anonymous reviewers for their constructive comments and suggestions.